\documentclass[10pt,twocolumn,letterpaper]{article}

\usepackage{cvpr}              
\usepackage{float}
\usepackage{pifont}

\definecolor{cvprblue}{rgb}{0.21,0.49,0.74}
\usepackage[pagebackref,breaklinks,colorlinks,allcolors=cvprblue]{hyperref}

\def\confName{CVPR}
\def\confYear{2027}

\title{Audio-Visual Rearrangement}

\author{First Author\\
Institution1\\
Institution1 address\\
{\tt\small firstauthor@i1.org}
\and
Second Author\\
Institution2\\
First line of institution2 address\\
{\tt\small secondauthor@i2.org}
}
  \author{
  Ruohao Guo$^{1}$,
  Jiatong Ye$^{1}$,
  Chenwei Dong$^{1}$,
  Jinxing Zhou$^{2}$,
  Dantong Niu$^{3}$,
  Guangyao Li$^{1}$,
  Zhenbo Li$^{1}$\\[0.5em]
  $^{1}$China Agricultural University\\
  $^{2}$Mohamed bin Zayed University of Artificial Intelligence\\
  $^{3}$University of California, Berkeley
  }

\begin{document}

\maketitle

\begin{abstract}
We introduce Audio-Visual Rearrangement (AVR), a novel task in which an agent hears a spoken request, retrieves a suitable object, and delivers it to the requesting user or an instruction-specified destination. AVR requires the agent to understand the spoken request and acoustically localize the speaker for delivery. To support this task, we construct a dataset covering three query types: explicit target specification, speaker-relative spatial references, and implicit user intent, spanning eight subtypes with diverse object selection and delivery requirements. Moreover, we propose a framework that integrates a motion-aware spatial audio module into an audio-visual large language model to encode spatial cues from the spoken request and update them as the agent moves. We further incorporate auxiliary supervision for speech understanding, spatial audio perception, and visual grounding through multitask instruction tuning to improve action prediction. Experiments establish baseline performance in both seen and unseen environments, laying the groundwork for future research on embodied intelligence and human-robot interaction. The code will be publicly available soon.
\end{abstract}
    
\section{Introduction}
\label{sec:intro}

\begin{figure}[t]
    \centering
    \includegraphics[width=\columnwidth]{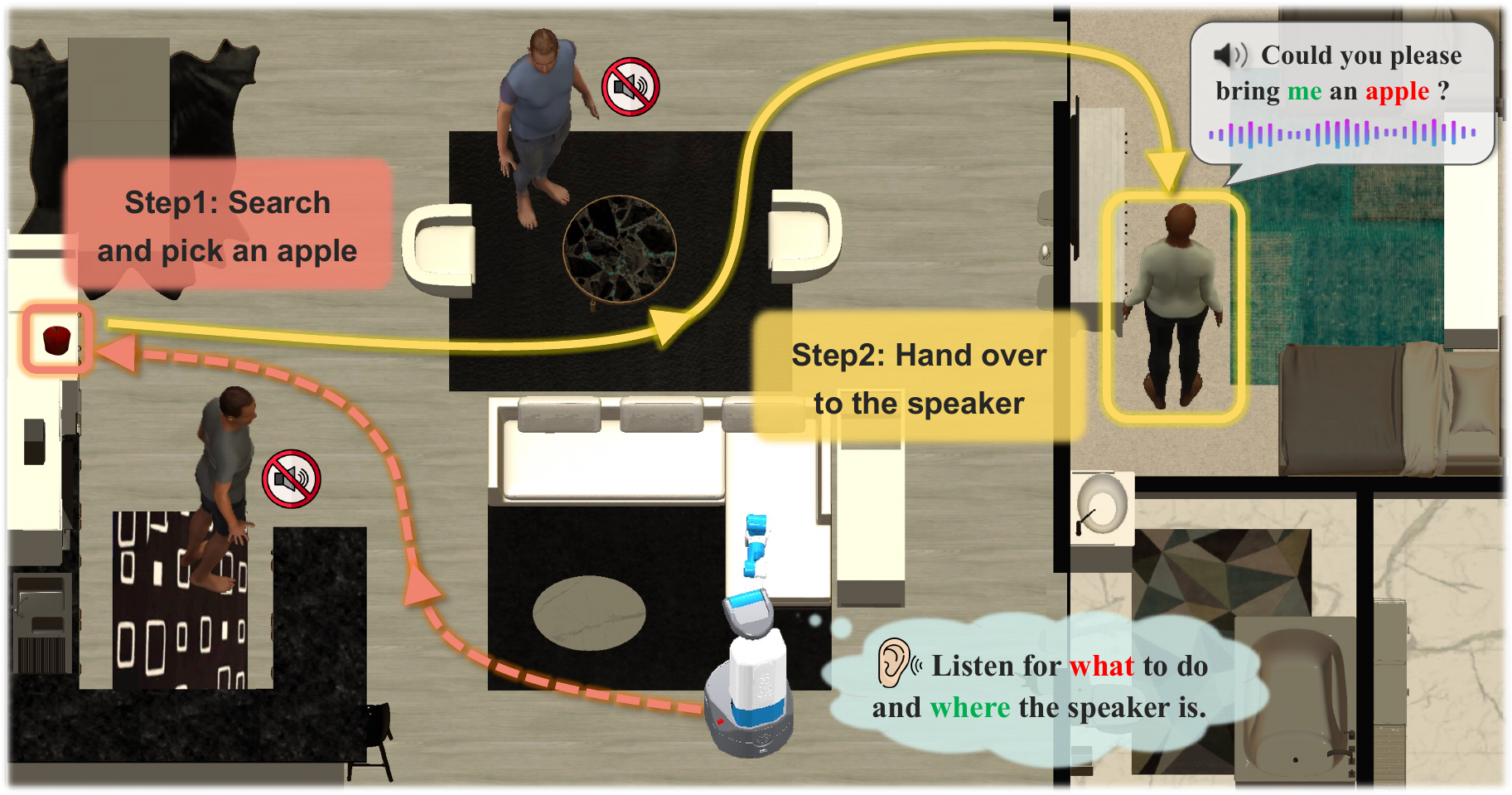}
    \caption{\textbf{Overview of Audio-Visual Rearrangement.} Given a spoken request, an agent must (1) search for and pick up a suitable object and (2) navigate to the requesting speaker and hand it over. This requires understanding what to retrieve from speech, visually grounding the target object, and acoustically localizing the speaker among multiple people, as well as navigation and manipulation.}
    \label{fig:task_overview}
\end{figure}

A household robot must reason about what it sees and hears to translate everyday human needs into physical actions. For example, when a user says, “I am thirsty,” the robot should infer the need for a drink, locate and retrieve a cup of water, and deliver it to the speaker. This interaction requires the robot to understand both what assistance is needed and where the intended recipient is located. Speech provides semantic cues about the user’s intent and spatial cues about the speaker, while visual observations support object localization and navigation. This setting extends audio-visual embodied research toward user-aware object rearrangement by bringing implicit intent understanding, speaker localization, and object manipulation into a unified task.

Recent advances in embodied rearrangement have enabled agents to restore rooms to target configurations~\cite{roomr}, infer commonsense object placements~\cite{tidee, tidybot}, and execute language-guided object retrieval and placement~\cite{alfred, homerobot, habitat3}. In parallel, audio-visual navigation has explored how agents can use visual observations and spatial acoustic cues to locate sound sources~\cite{soundspaces, semantic_audio_visual_navigation, rila, savnce}. These research directions address complementary aspects of household assistance, but typically study them separately. Audio-visual navigation focuses on reaching a sound source, with limited attention to retrieving an object and delivering it to the speaker. Embodied rearrangement benchmarks primarily specify tasks through text instructions or visual goal states, without requiring agents to locate the requester using speech. In everyday assistance, however, these abilities are closely connected: the same utterance conveys what the user needs and provides acoustic cues about where the user is. This calls for agents to combine speech understanding and speaker localization with visual perception to retrieve a suitable object and deliver it to the intended person—a setting that remains underexplored in existing benchmarks.

In light of these limitations, we introduce Audio-Visual Rearrangement (AVR), a novel task in which an embodied agent follows a spoken request to retrieve a suitable object and deliver it to the requesting user or place it on a designated support surface (Figure~\ref{fig:task_overview}). In unfamiliar indoor environments with multiple people, AVR requires an agent to infer which object fulfills a spoken request and locate the requesting speaker. It introduces closely related challenges in intent understanding and spatial reasoning: the agent needs to infer which object fulfills the request and locate the speaker, who may serve as both the intended recipient and a spatial reference for selecting the object or delivery location. Since the request is spoken only once, the agent also needs to remember the speaker’s location and update its position relative to the speaker throughout exploration and manipulation. AVR therefore offers a task setting for studying how agents integrate spoken intent, spatial perception, and physical interaction for household assistance.

To tackle AVR, we develop an omni-modal framework based on Qwen2.5-Omni~\cite{qwen25omni} that integrates spoken language understanding, motion-aware spatial hearing, and closed-loop visual skill prediction. The framework uses the initial spoken request to infer both the speaker's intent and location. Specifically, a Motion-Aware Spatial (MAS) Audio Encoder extracts directional and distance cues from the binaural utterance. As the agent moves, a Motion Adapter updates these spatial features based on the agent's pose relative to its initial listening pose, maintaining speaker-relative information without repeated speech. A spatial projector maps the adapted features into tokens, which the language model integrates with the instruction, visual observations, and execution feedback to predict navigation and manipulation skills. This design links object-directed exploration to subsequent navigation toward the speaker. We train the framework with a three-stage curriculum: spatial audio pretraining develops localization capabilities, motion adaptation learns to update spatial features as the agent moves, and multitask instruction tuning aligns these representations with instruction understanding and action prediction.

To support the study of AVR, we build a simulation benchmark featuring spoken requests, binaural spatial audio, and multiple people in furnished indoor environments. The benchmark covers explicit-target, spatial-reference, and implicit-intent requests, with 2,000 episodes for training and validation and 400 evaluation episodes across 49 scenes. Our method achieves overall success rates of 20.5\% on seen scenes and 17.0\% on unseen scenes, outperforming the evaluated baselines across all three request types. Ablation studies show improvements in overall task success with spatial audio encoding and motion adaptation, with further gains from the combined use of multitask supervision and response-loss weighting. Additional analyses characterize successful executions across initial speaker distances and performance under concurrent environmental sounds. These results highlight the potential of integrating spoken semantics, spatial hearing, and physical interaction for household assistance, while identifying acoustic interference as a key remaining challenge.

Our main contributions are summarized as follows:

1) We introduce Audio-Visual Rearrangement (AVR), a task in which an agent follows a spoken request to retrieve a suitable object and deliver it to the requesting speaker among multiple people or an instruction-specified destination, jointly requiring spoken intent understanding, speaker localization, and object manipulation.

2) We develop an omni-modal baseline framework that extracts spatial features from a single initial utterance and conditions them on the agent's relative motion, integrating these features with spoken intent and visual observations to predict executable navigation and manipulation skills.

3) We establish a simulation benchmark covering three request types across seen and unseen scenes. Experiments compare baseline systems, ablations assess the contributions of the framework's components and training configuration, and additional analyses characterize its sensitivity to concurrent environmental sounds.

\section{Related Work}
\label{sec:related_work}

\subsection{Audio-Visual Navigation}
Audio-visual navigation~\cite{soundspaces,audio_visual_waypoints} requires an agent to combine visual observations with acoustic cues to locate and reach a sound-emitting object in an unfamiliar environment. To support this task, Chen et al.~\cite{soundspaces} introduce SoundSpaces, a platform for geometrical acoustic simulation in 3D environments. Semantic audio-visual navigation~\cite{semantic_audio_visual_navigation,rila} extends this task to semantically meaningful and intermittent object sounds, where an agent must infer the target object's category and location and use past observations to reach it even after the sound stops. SAVN-CE~\cite{savnce} brings this task into continuous environments using the on-the-fly acoustic rendering of SoundSpaces 2.0~\cite{soundspaces2}, allowing agents to move freely through 3D scenes while receiving spatially and temporally coherent audio-visual observations. Unlike these navigation tasks, our task requires an agent to follow a spoken instruction to retrieve and deliver an object to the speaker or an instruction-specified location, using both the speech semantics and acoustic localization.

\subsection{Embodied Rearrangement}
Embodied rearrangement~\cite{rearrangement} aims to transform an environment into a desired configuration through navigation and physical interaction. In visual room rearrangement~\cite{roomr}, an agent restores object positions and states to a previously observed room configuration. Other studies~\cite{tidee,tidybot,alfred,homerobot} address household rearrangement and mobile manipulation without an explicit visual reference for the goal state. These tasks involve either inferring appropriate object placements from commonsense knowledge or user preferences~\cite{tidee,tidybot}, or following natural-language instructions that specify the desired outcomes~\cite{alfred,homerobot,teach}. The former emphasizes deciding where objects belong and relocating them accordingly, whereas the latter focuses on grounding instructions in visual observations to guide object search, navigation, and manipulation. Beyond single-agent execution, human--robot collaboration~\cite{habitat3,partnr} introduces coordination with a partner in a shared environment. Habitat 3.0~\cite{habitat3} supports this setting through humanoid simulation and human-in-the-loop interaction. Its social rearrangement task pairs a robot with a humanoid partner to efficiently move objects to designated locations. Compared to text-based instruction-following tasks, our task involves not only understanding a spoken request but also locating the speaker and delivering the requested object to them, more closely reflecting how people seek robotic assistance in everyday life.

\section{Audio-Visual Rearrangement}
\label{sec:task}

We introduce Audio-Visual Rearrangement (AVR), a novel task in which an embodied agent follows a spoken request to retrieve a suitable object and deliver it to the requesting user or place it on an instruction-specified support surface. In this task, the spoken request provides both semantic cues about the user's intent (e.g., ``bring me a cup'') and spatial cues about the user's location. This setting poses new challenges for embodied rearrangement~\cite{rearrangement}. The agent needs to ground the intent conveyed by speech in visually observed objects and use spatial audio cues to identify and localize the requesting speaker among multiple people.

\subsection{Simulation Environment} 
\label{sec:sim_env}
We build our simulation environment on Habitat 3.0~\cite{habitat3}, using furnished indoor scenes from HSSD~\cite{hssd}. Each environment contains multiple humanoid avatars from Habitat's human assets and an embodied agent modeled as a Fetch mobile manipulator with an egocentric camera. One avatar is designated as the requesting user, while the others serve as distractors. We synthesize speech for each request using Qwen3-TTS~\cite{qwen3tts} and associate the resulting waveform with the requesting user. To simulate the audio received by the agent, we use the SoundSpaces 2.0~\cite{soundspaces2} acoustic renderer integrated into Habitat-Sim. Specifically, we position the sound source at the requesting user's head and align the listener's position and orientation with the agent's camera. We then convolve the synthesized speech waveform with the simulated binaural room impulse responses to generate spatialized speech, coupling the request's semantic content with acoustic cues to the speaker's location.

\subsection{Task Taxonomy and Formulation}
\label{sec:taxonomy}
We define three query types with eight subtypes. \textbf{Explicit Target} queries directly name the object category and include \emph{Category} queries, which accept any eligible instance (e.g., ``bring me a cup''), and \emph{Instance} queries, which identify a specific instance by its room or supporting furniture (e.g., ``bring me the cup on the kitchen counter''). \textbf{Spatial Reference} queries specify a spatial relation to the requesting user and include \emph{Object}, which requires retrieving the nearest eligible object (e.g., ``bring me the cup closest to me''), and \emph{Destination}, which requires placing an object on the nearest eligible support (e.g., ``put the cup on the table closest to me''). These queries require identifying the requesting speaker as the spatial reference and comparing candidate locations relative to that person. \textbf{Implicit Intent} queries express a need or intended activity without naming a target object, requiring the agent to infer a suitable category. They cover four domains: \emph{Food \& Drink}, including meal preparation (e.g., ``I'm thirsty''); \emph{Home}, covering furnishing and decoration (e.g., ``bring me something to decorate this space''); \emph{Work}, covering work, study, and organization (e.g., ``I need to get ready for a study session''); and \emph{Leisure}, covering play and entertainment (e.g., ``I'm bored and want something to play with''). Each Implicit Intent query is associated with a set of acceptable object categories, allowing different objects to satisfy the same request under the benchmark's category-based criterion.

Each episode specifies an indoor scene, a spoken request from a designated user, and the acceptable objects and delivery goal. The agent and avatars are initialized at randomly sampled, collision-free locations in navigable areas of the scene. The requesting user utters the request only once at the beginning of the episode, with no further speech during task execution. An episode is successful only if the agent retrieves an acceptable object and completes the required delivery within the episode budget. For delivery to the user, the agent must hold the object, be within handover range of the requesting user, and successfully complete the handover action for that object. Approaching another avatar does not satisfy this condition. For the Destination subtype of Spatial Reference, the agent must release the object onto the eligible support nearest to the requesting user, where it must remain stably supported at the end of the episode. Thus, success jointly requires correct object selection and delivery to the correct recipient or support.

\section{Method}
\label{sec:method}

\begin{figure*}[t]
  \centering
  \includegraphics[width=\linewidth]{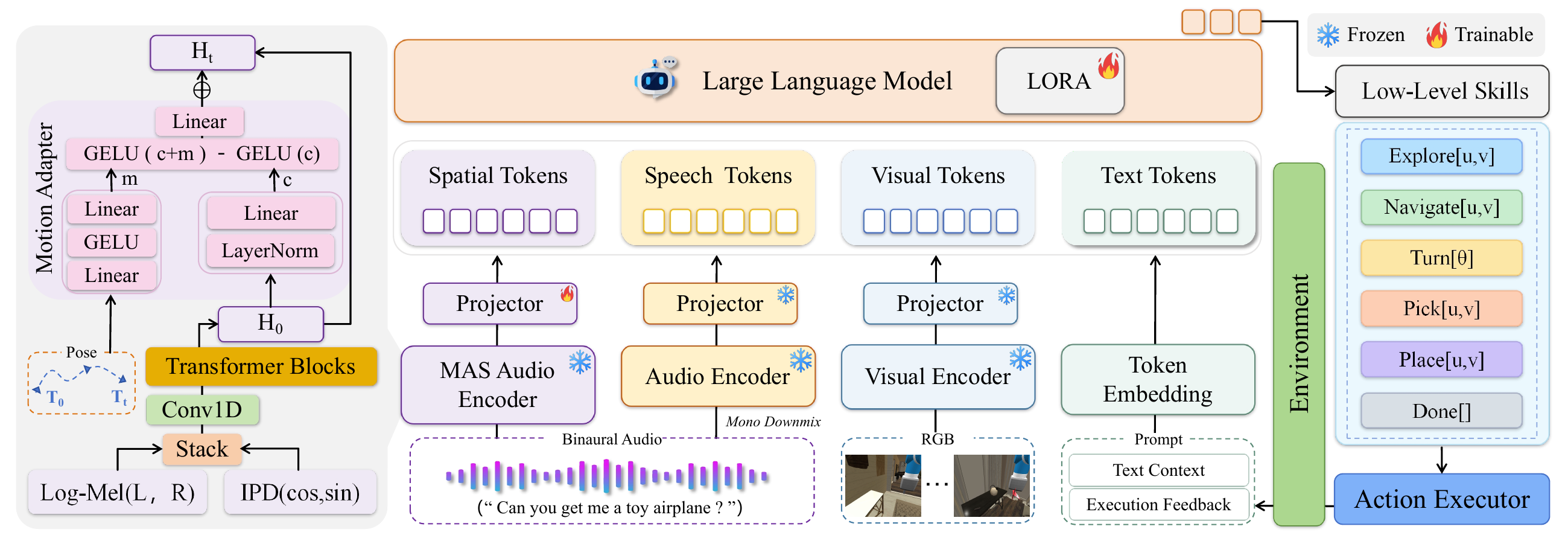}
  \caption{\textbf{Overview of our baseline framework.} Details of the Motion-Aware Spatial (MAS) Audio Encoder are shown in the left panel. The original audio encoder preserves speech understanding, while the parallel MAS Audio Encoder extracts binaural spatial cues and updates them according to the agent's motion. A lightweight projector maps these features into the language model's token space, where they are combined with speech, visual, and textual tokens to predict executable skills. The action executor interacts with the environment and returns execution feedback for closed-loop decision making.}
  \label{fig:framework}
\end{figure*}

We introduce a baseline framework for audio-visual rearrangement that combines speech understanding and motion-aware spatial audio perception with visual grounding for closed-loop skill prediction. We first present the model architecture, highlighting the proposed Motion-Aware Spatial (MAS) Audio Encoder (Sec.~\ref{sec:arch}), and then describe the three-stage training pipeline for spatial perception, motion adaptation, and embodied decision making (Sec.~\ref{sec:training}).

\subsection{Model Architecture}
\label{sec:arch}

Figure~\ref{fig:framework} shows the overall architecture. We build on Qwen-Omni model~\cite{qwen25omni}, retaining its visual encoder, monaural audio encoder, and Thinker language-model backbone. We introduce an MAS Audio Encoder that extracts spatial cues from binaural audio and updates them according to the agent's motion through an integrated Motion Adapter. A lightweight projector maps its outputs into spatial tokens for the language model. Complementing the speech content captured by the monaural branch, these tokens represent the speaker's direction and distance relative to the agent. The language model combines spatial, speech, visual, and textual tokens to predict executable skills.

\noindent\textbf{Spatial audio feature encoding.}
To capture acoustic cues about the speaker's direction and distance, the MAS Audio Encoder first encodes binaural audio into spatial features. Given a binaural recording, we construct the input features as $\operatorname{Stack}[M_L, M_R, \mathcal{B}(\cos\Delta\phi), \mathcal{B}(\sin\Delta\phi)]$, where $M_L$ and $M_R$ are the left- and right-ear log-mel features, $\Delta\phi$ denotes the interaural phase difference, and $\mathcal{B}$ denotes magnitude-weighted mel-band aggregation. The frontend preserves absolute signal levels and interaural differences to retain acoustic localization cues. These features are processed by temporal convolutions and a Transformer backbone, followed by adaptive average pooling to compress the output sequence into 16 spatial feature vectors $H_0$. They represent the acoustic observation at the initial receiver pose and serve as the reference for subsequent motion adaptation.

\noindent\textbf{Motion adaptation.}
As the agent moves from the recording pose $T_0$ to the current receiver pose $T_t$, the speaker's position relative to the receiver changes. Assuming that the speaker remains stationary, the Motion Adapter updates the initial acoustic features $H_0$ using the relative rotation and translation derived from $(T_0,T_t)$, producing spatial features $H_t$ conditioned on the current receiver pose. As illustrated in the left panel of Fig.~\ref{fig:framework}, a motion branch encodes the relative pose into a motion embedding $m$, while an acoustic branch applies LayerNorm and a linear projection to $H_0$ to obtain a conditioning feature $c$. The two branches interact through the residual bottleneck $\operatorname{GELU}(c+m)-\operatorname{GELU}(c)$, whose output is linearly projected and added to $H_0$ to produce $H_t$. Each $H_t$ is computed directly from $H_0$ rather than recursively from previous features, maintaining a motion-conditioned spatial representation of the initial utterance without requiring the speaker to repeat the request.

\noindent\textbf{Skill prediction.}
To translate the spoken request into executable skills, the language model combines spatial audio features, speech content, visual observations, and textual context. A trainable MLP projector maps the adapted spatial features $H_t$ into spatial tokens in the language model's embedding space. The monaural branch encodes speech content from channel-averaged binaural audio, while visual observations provide scene and object information. Textual context includes the task prompt, a retained textual representation of the initial request, and execution feedback. At subsequent steps, the retained request and updated spatial features support decision making without repeated speech, while current observations and selected historical views support object grounding and speaker identification. At each decision step, the language model predicts a skill $a_t$ from \texttt{Explore[$u,v$]}, \texttt{Navigate[$u,v$]}, \texttt{Turn[$\theta$]}, \texttt{Pick[$u,v$]}, \texttt{Place[$u,v$]}, and \texttt{Done[]}. For coordinate-based skills, $(u,v)$ specifies a target location in the current egocentric image, normalized to the range $[0,100]$, while $\theta$ specifies a discrete rotation angle. The \texttt{Place} skill delivers the held object to the designated recipient or support, and \texttt{Done[]} terminates execution. A low-level controller executes skills using simulator depth, camera poses, navmesh, and instance observations, returning observations and execution feedback for closed-loop prediction.

\subsection{Training Pipeline}
\label{sec:training}
We adopt a three-stage training pipeline: spatial audio pretraining, motion adaptation pretraining, and multitask instruction tuning. In Stage 1, we pretrain the binaural feature encoding pathway of the MAS Audio Encoder to predict the speaker's direction and distance relative to the receiver. In Stage 2, we freeze the pretrained acoustic encoding pathway and train the Motion Adapter to update the initial spatial features according to the receiver's relative motion. We retain the localization objective from Stage 1, with target direction and distance expressed in the current receiver frame. In Stage 3, we jointly optimize the spatial projector and language model LoRA parameters~\cite{lora}, while keeping the MAS Audio Encoder, the original audio and visual encoders, and the base language model weights frozen. The training mixture comprises ten instruction--response tasks: speech transcription, source localization, target category extraction, visual target grounding, instruction--target matching, speaker grounding in current observations, speaker grounding in historical views, scene/furniture visibility prediction, main policy prediction, and delivery policy prediction. These tasks provide complementary supervision for auditory understanding, visual perception, audio-visual spatial association, and embodied decision making, supporting skill prediction and delivery to the requesting user or an instruction-specified destination.

\section{Experiments}
\label{sec:experiments}

\begin{table*}[h]
\centering
\caption{Evaluation on the audio-visual rearrangement benchmark across seen and unseen environments.}
\label{tab:main_avr}
\begin{tabular}{l|rrrrr|rrrrr}
\hline
& \multicolumn{5}{c|}{Seen} & \multicolumn{5}{c}{Unseen} \\
& ESR $\uparrow$ & RSR $\uparrow$ & ISR $\uparrow$ & OSR $\uparrow$ & Steps $\downarrow$ & ESR $\uparrow$ & RSR $\uparrow$ & ISR $\uparrow$ & OSR $\uparrow$ & Steps $\downarrow$ \\
\hline
Random & 0.0 & 0.0 & 0.0 & 0.0 & 5000.0 & 0.0 & 0.0 & 0.0 & 0.0 & 5000.0 \\
Omni-VLM & 0.0 & 0.0 & 0.0 & 0.0 & 4927.6 & 0.0 & 0.0 & 0.0 & 0.0 & 4921.3 \\
Modular Language Agent & 11.3 & 1.7 & 3.3 & 6.0 & 3729.7 & 12.5 & 1.7 & 5.0 & 7.0 & 3851.2 \\
LLM Planner & 20.0 & 0.0 & 15.0 & 12.5 & 2973.9 & 17.5 & 1.7 & 13.3 & 11.5 & 2867.8 \\
Ours & \textbf{25.0} & \textbf{16.7} & \textbf{18.3} & \textbf{20.5} & \textbf{2568.5} & \textbf{18.8} & \textbf{15.0} & \textbf{16.7} & \textbf{17.0} & \textbf{2685.5} \\
\hline
\end{tabular}
\end{table*}

\subsection{Experimental Setting}
\label{sec:setting}

\begin{figure}[t]
    \centering
    \includegraphics[width=\columnwidth]{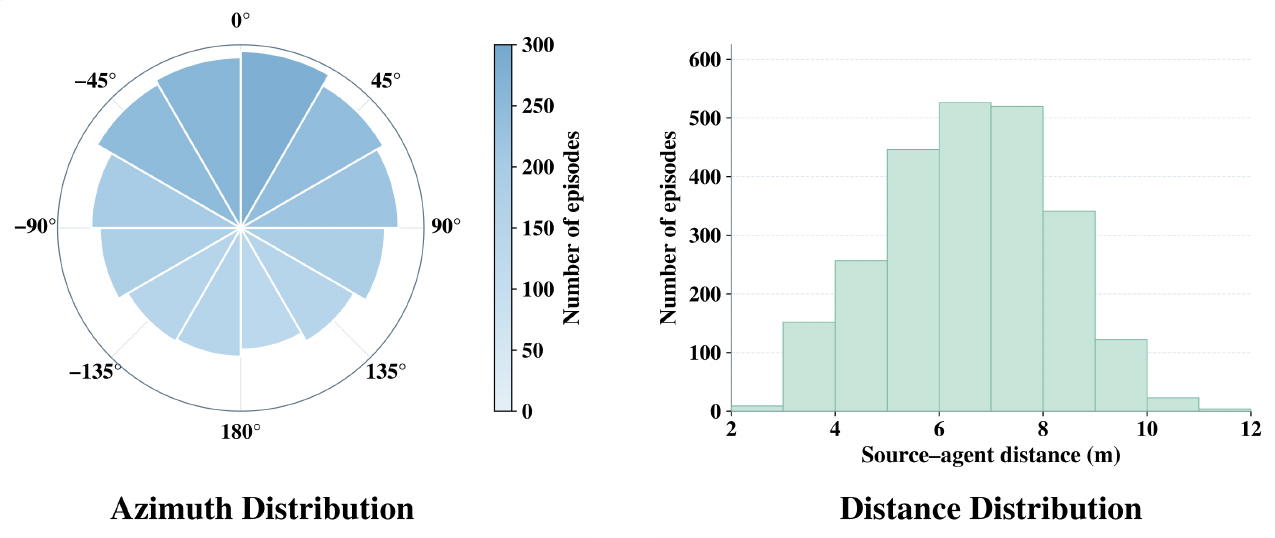}
    \caption{\textbf{Distributions of initial speaker azimuths and source-agent Euclidean distances across AVR episodes.} \textbf{Left}: Polar histogram showing the spatial distribution of initial speaker azimuths relative to the agent. \textbf{Right}: Histogram illustrating the distribution of source-agent distances, spanning from 2 to 12 meters.}
    \label{fig:task_setting}
\end{figure}

\noindent\textbf{Dataset Construction.}
We construct instruction-tuning data from 2,000 base AVR episodes across 38 development scenes, split by scene into 1,790 training episodes in 34 scenes and 210 validation episodes in four scenes. A rule-based expert uses privileged simulator object and speaker states to select targets and generate current-image pixel actions. We retain all decisions from accepted base trajectories and supplement underrepresented actions with verified successful examples and resampling. The ten-task mixture contains 150k training and 13k validation examples. Evaluation uses 400 additional episodes: 200 in the 38 development scenes (seen) and 200 in 11 scenes outside this pool (unseen). Here, ``seen'' includes both instruction-tuning training and validation scenes. Both evaluation subsets follow Sec.~\ref{sec:taxonomy}, with explicit, relational, and implicit requests in a 4:3:3 ratio. Figure~\ref{fig:task_setting} shows initial speaker azimuths and source-to-agent Euclidean distances across the 2,400 episodes. Azimuths cover the full angular range; distances concentrate at 5--9~m and peak at 6--8~m.

\noindent\textbf{Metrics.}
We report success rate (SR) and mean simulator steps (Steps). Success requires satisfying all object and destination constraints specified by the request. ESR, RSR, and ISR denote SR for explicit, relational, and implicit requests, respectively, while OSR denotes overall SR across all request types. Steps is the mean execution length across all episodes, including failures and timeouts, under a 5,000-step budget per episode. Since early termination also contributes to this average, Steps reflects overall execution length rather than efficiency on successfully completed tasks and should be interpreted alongside SR. A lower Steps value alone therefore does not necessarily indicate better task performance. To distinguish target selection from delivery, we additionally report target selection accuracy (TSA) and conditional delivery success rate (CDSR). TSA measures whether the acquired object satisfies the request, computed as the fraction of episodes with a correct target acquired among those with any object acquired. Episodes without object acquisition are excluded from TSA. CDSR measures delivery success after correct target acquisition, computed as the fraction of fully successful episodes among those with a correct target acquired. Delivery includes handovers to the correct recipient and stable placements on the designated support.

\noindent\textbf{Baselines.}
We compare our method against four baselines under the same task settings and evaluation protocol. (1) Random: A random policy that uniformly samples six high-level action types: exploration, navigation, turning, picking, surface placement, and handover, with randomly sampled action parameters. It stops when an oracle detects task completion (perfect stopping). (2) Omni-VLM: A pretrained Qwen2.5-Omni model that predicts actions without task-specific fine-tuning. It uses the same observation inputs, history window, and action interface as our method. (3) Modular Language Agent: A modular agent that uses Qwen2.5-7B-Instruct for planning, Whisper for speech recognition, and a learned audio localization module for sound-source estimation. GroundingDINO and SAM process RGB-D observations to construct an online semantic map. Starting without a scene map, the agent selects high-level actions based on the transcribed instruction and the observed map. (4) 
LLM Planner: An LLM-based planner initialized with a ground-truth scene graph that encodes room membership and object--support relations and is updated during execution. Whisper provides instruction transcriptions, while an audio localization module estimates the sound-source location.

\noindent\textbf{Implementation Details.}
Audio is sampled at 16~kHz. The spatial encoder has 32 Transformer layers with hidden size 1280 and outputs 2048-dimensional features. A motion adapter with bottleneck size 128 updates them using simulator-provided relative receiver poses before projection into 16 spatial tokens of dimension 3584 for Qwen2.5-Omni-7B. Stage~1 uses a scene-disjoint train/validation split of 33/2 scenes. Stage~2 uses a scene-disjoint train/validation split of 34/4 scenes. Both splits exclude all 11 unseen evaluation scenes. Stage~1 trains for 10 epochs with cosine direction and smooth $L_1$ log-distance losses, backbone/head learning rates of $10^{-5}/10^{-4}$, and weight decay 0.01. Stage~2 trains the motion adapter and localization head on frozen features using AdamW, learning rates of $10^{-3}/10^{-4}$, batch size 256, and no weight decay. Cosine annealing uses a minimum learning rate of $10^{-5}$; we select the epoch-40 checkpoint. Stage~3 trains only the projector and LoRA parameters on the ten-task mixture. Response-token cross-entropy weights are 2.0 for action names and arguments, 0.25 for \texttt{Assigned!}, and 1.0 otherwise. AdamW uses learning rate $10^{-5}$, weight decay 0.1, and cosine decay with 3\% warmup. LoRA uses rank 16, $\alpha=32$, and dropout 0.05. Stage~3 runs for one epoch on six A100 GPUs in BF16, with batch size 1 per GPU and eight accumulation steps (effective batch size 48).

\begin{table}[h]
\centering
\caption{Success rates (\%) of our method across eight AVR query subtypes on the seen and unseen splits.}
\label{tab:avr_query_subtypes}
\begin{tabular}{l|l|rr}
\hline
 & Subtype & Seen & Unseen \\
\hline
Explicit Target
& Category      & 30.0 & 23.3 \\
& Instance      & 22.0 & 16.0 \\
\hline
Spatial Reference
& Object        & 30.0 & 26.7 \\
& Destination   &  3.3 &  3.3 \\
\hline
Implicit Intent
& Food \& Drink & 26.7 & 20.0 \\
& Home          & 13.3 & 13.3 \\
& Work          & 20.0 & 20.0 \\
& Leisure       & 13.3 & 13.3 \\
\hline
\multicolumn{2}{l}{Overall}
& 20.5 & 17.0 \\
\hline
\end{tabular}
\end{table}

\subsection{Comparison with Baselines}
\label{sec:mainres}

Table~\ref{tab:main_avr} compares baseline performance on AVR. The zero-shot Omni-VLM policy completes no tasks, suggesting that pretrained omni-modal capabilities alone are insufficient for reliable execution under this protocol. The Modular Language Agent achieves nonzero success using our sound-source localizer. LLM Planner performs better overall with ground-truth room membership and object--support relations, although its low RSR may partly reflect missing speaker-relative spatial information in the scene graph. Our method achieves 20.5\% and 17.0\% OSR on seen and unseen scenes, exceeding LLM Planner by 8.0 and 5.5 percentage points, respectively. It also achieves the highest SR across all three request types on both splits and the lowest mean execution lengths of 2568.5 and 2685.5 steps. These results show that our framework achieves higher task success than the evaluated baselines in retrieving suitable objects and delivering them to the requesting speaker or an instruction-specified destination.

Table~\ref{tab:avr_query_subtypes} breaks down performance by request subtype. Retrieving any object from a named category is easier than retrieving a location-specific object on both splits. Speaker-relative retrieval achieves 30.0\%/26.7\% SR on seen/unseen scenes, while nearest-support placement reaches only 3.3\% on both, highlighting speaker-relative placement as a remaining challenge. For implicit requests, food and drink achieves 26.7\%/20.0\%, work and study 20.0\% on both splits, and home decoration and leisure 13.3\% each, showing success across all four domains but varying performance across user needs.

\begin{figure*}[h]
\centering
\includegraphics[width=\linewidth]{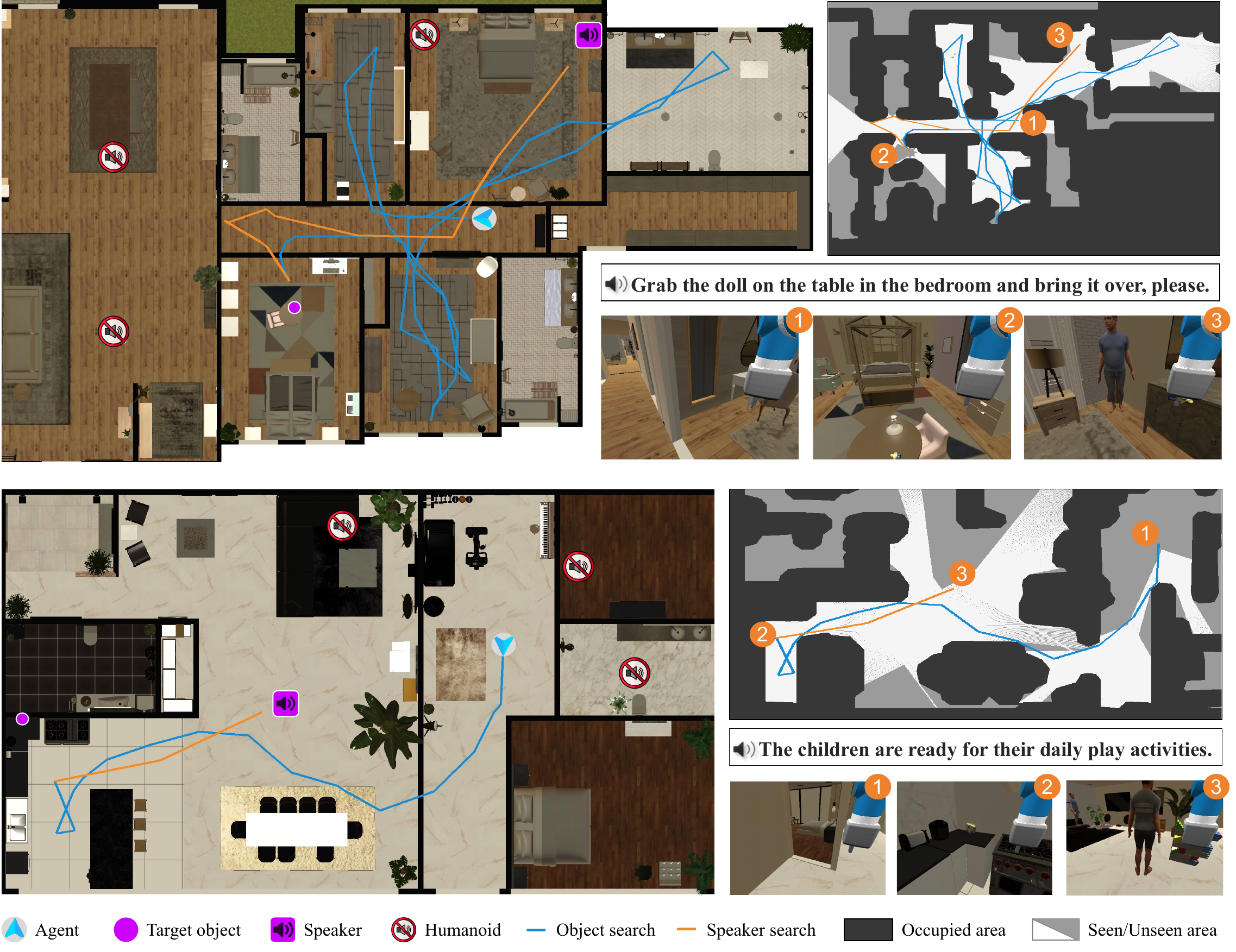}
\caption{\textbf{AVR trajectories in unseen environments.} The top and bottom examples correspond to explicit and implicit requests. Each example shows a top-down scene view, an occupancy map, and egocentric observations during exploration (1), target acquisition (2), and approach to the speaker (3). Blue and orange paths indicate object search and sound-source search. Paths connect recorded decision locations along the navigation mesh. Selected views include a simulator-rendered snapshot for illustration.}
\label{fig:qualitative}
\end{figure*}

\subsection{Qualitative Results}
\label{sec:qualitative}

Figure~\ref{fig:qualitative} presents two AVR trajectories in unseen environments, with blue and orange paths indicating object search and speaker search, respectively. In the explicit-request example (top), the speaker asks for a doll on a bedroom table. The specified category and location constrain the valid target, requiring extended exploration before retrieval. The agent then navigates toward the requesting speaker while carrying the doll. In the implicit-request example (bottom), the speaker describes children's play activities without specifying an object. The agent infers that a toy would satisfy this need. Since multiple toys are available in the room, any suitable one fulfills the request, allowing a shorter object search. Thus, the two requests impose different constraints on object selection. Both scenes contain multiple people, requiring the agent to distinguish the requesting speaker from other individuals. Since the request is spoken only once, the agent retains and updates speaker-relative spatial information throughout exploration and retrieval. The numbered observations show object acquisition and the subsequent approach to the speaker. Together, these examples illustrate how request specificity affects exploration and how our framework connects spoken intent with object retrieval and speaker-directed navigation.

\subsection{Ablation Studies}
\label{sec:ablation}

Table~\ref{tab:ablation_avr} examines the contributions of spatial audio encoding, motion adaptation, and the combined training strategy. The first three configurations share the same policy training examples and response cross-entropy loss. Adding the frozen pretrained spatial audio encoder to the monaural baseline increases OSR from 3.5\% to 7.5\%, while TSA improves from 34.4\% to 46.5\% and CDSR from 15.9\% to 25.0\%. These improvements suggest that spatial audio provides useful information for object retrieval and delivery, consistent with the speaker's role as both a spatial reference and a potential recipient. Incorporating the frozen pretrained motion adapter further increases OSR to 11.0\% and CDSR to 34.4\%. Although TSA decreases slightly from 46.5\% to 43.5\%, overall task success continues to improve, suggesting that initial spatial features become more useful for task execution when conditioned on the agent's relative motion. The accompanying increase in CDSR is consistent with the motivation to preserve speaker-relative information throughout exploration and manipulation, supporting subsequent delivery after the agent has moved away from its initial listening position.

\begin{table}[h]
\centering
\caption{Ablation study on the unseen split. Components are added cumulatively to the monaural baseline. Ours further incorporates multitask supervision and response-loss weighting.}
\label{tab:ablation_avr}
\small
\setlength{\tabcolsep}{5pt}
\resizebox{\columnwidth}{!}{%
\begin{tabular}{l|rrrr}
\hline

& OSR $\uparrow$
& TSA $\uparrow$
& CDSR $\uparrow$
& Steps $\downarrow$ \\
\hline
Monaural & 3.5 & 34.4 & 15.9 & 2621.9 \\
+ Spatial Audio & 7.5 & 46.5 & 25.0 & 2926.4 \\
+ Motion Adapter & 11.0 & 43.5 & 34.4 & 2752.4 \\
Ours & 17.0 & 47.6 & 49.3 & 2685.5 \\
\hline
\end{tabular}%
}
\end{table}

Our full model combines a ten-task mixture of policy learning and auxiliary supervision with response-loss weighting, achieving the highest OSR, TSA, and CDSR among the evaluated configurations: 17.0\%, 47.6\%, and 49.3\%, respectively. The auxiliary tasks supervise instruction understanding, visual grounding, and speaker localization, while response-loss weighting emphasizes action names and arguments. Together, these changes improve OSR by 6.0 percentage points over the motion-adapted configuration, with gains in both TSA and CDSR. As a complementary spatial diagnostic, the localization head applied to motion-adapted features achieves a wrapped azimuth MAE of $19.41^\circ$ in the current receiver frame and a horizontal-distance MAE of 1.058~m on the Stage~2 validation subset. These results provide complementary evidence of spatial localization accuracy and the benefits of the combined training strategy for task completion.

\begin{table}[h]
\centering
\caption{Robustness to spatial distractors on the unseen split. }
\label{tab:distractor_avr}
\small
\setlength{\tabcolsep}{5pt}
\begin{tabular}{l|rrrr}
\hline

& OSR $\uparrow$
& TSA $\uparrow$
& CDSR $\uparrow$
& Steps $\downarrow$ \\
\hline
Clean & 17.0 & 47.6 & 49.3 & 2685.5 \\
Distractor & 6.5 & 37.2 & 27.1 & 2738.1 \\
\hline
\end{tabular}%
\end{table}
\subsection{Robustness to Spatial Distractors}
\label{sec:robustness}
To evaluate robustness against acoustic interference, we introduce a concurrent environmental sound at a navigable location at least 2~m from the speaker and 1~m from the agent, with active power matched to the target signal. On the same 200 unseen episodes, distractors reduce OSR from 17.0\% to 6.5\%, TSA from 47.6\% to 37.2\%, and CDSR from 49.3\% to 27.1\%, while Steps increase to 2738.1 (Table~\ref{tab:distractor_avr}). The decline in TSA indicates poorer target selection among episodes with successful object acquisition, while the lower CDSR shows that delivery remains difficult even after acquiring a correct target. Acoustic interference therefore affects both target acquisition and subsequent delivery, highlighting a limitation of the current framework in multi-source environments.

\begin{figure}[H]
\centering
\includegraphics[width=\linewidth]{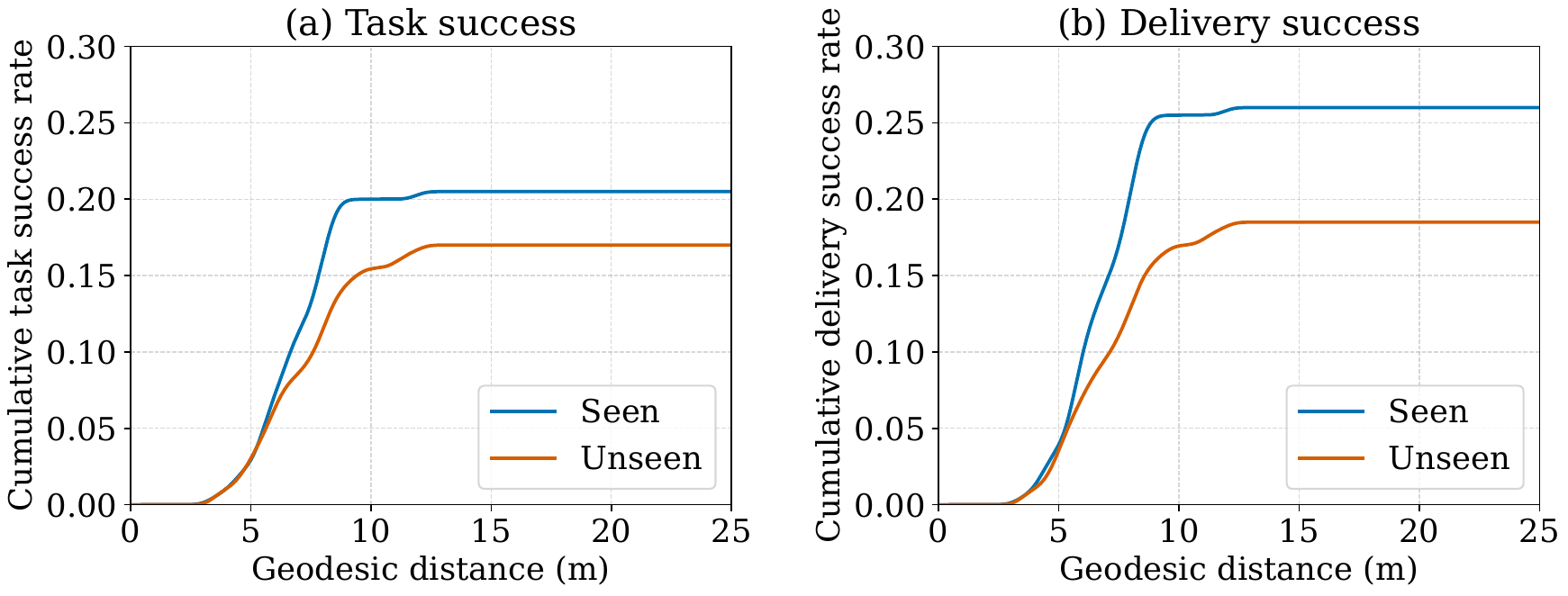}
\caption{\textbf{Cumulative success across initial Geodesic distances.} (a) Complete AVR task success. (b) Handover to the correct recipient. Blue and orange denote seen and unseen episodes. Distances are computed on the static-scene navigation mesh.}
\label{fig:distance}
\end{figure}

\subsection{Effect of Initial Distance}
\label{sec:distance}

Our dataset initializes the agent and speaker at Geodesic distances of 2--25~m. Figure~\ref{fig:distance} reports cumulative success against initial geodesic distance, which accounts for obstacles and room layouts. Within the evaluated distance distribution, both task success and recipient-handover success approach their overall rates around 10--12~m, with few additional successes at larger distances. This pattern occurs on both seen and unseen splits and is consistent with the increased demands of distant-speaker localization and navigation. Greater separation and intervening obstacles can weaken direct-path speech and make spatial cues less reliable, while longer navigation routes introduce more opportunities for execution errors. Since the request is spoken only once, the agent must also maintain speaker-relative information throughout these longer trajectories.

\section{Conclusion}
\label{sec:conclusion}

We introduce AVR, a novel task that connects spoken requests with object retrieval and delivery in multi-person indoor environments. Our omni-modal framework combines an MAS Audio Encoder with a Motion Adapter to extract spatial cues from a single initial utterance and update them as the agent moves. These cues are integrated with spoken intent and visual observations to support closed-loop navigation and manipulation. Experiments demonstrate that our framework outperforms the evaluated baselines, with particularly notable gains on speaker-relative requests. The results also reveal remaining challenges in reliable task execution in unseen environments and robustness to concurrent acoustic interference. Future work will investigate multi-source audio understanding and transfer to real-world mobile manipulators, extending this framework toward practical household assistance that connects spoken human needs with physical actions.

\section*{LLM Usage Statement}
The authors used large language models solely to improve the language and clarity of the manuscript. These tools were not used for research ideation, experimental design, data analysis, or drawing scientific conclusions. The authors reviewed and verified all revisions and take full responsibility for the content of this paper.
{
    \small
    \bibliographystyle{ieeenat_fullname}
    \bibliography{main}
}


\end{document}